# MetaSSC: Enhancing 3D Semantic Scene Completion for Autonomous Driving through Meta-Learning and Long-sequence Modeling

Yansong Qu[a], Zixuan Xu[b], Zilin Huang[c], Zihao Sheng[c], Sikai Chen[c]*, Tiantian Chen[b]*

[a] *Lyles School of Civil Engineering, Purdue University, West Lafayette, 47907, USA*

[b] *Cho Chun Shik Graduate School of Mobility, KAIST, Daejeon, 34051, South Korea*

[c] *Department of Civil and Environmental Engineering, University of Wisconsin-Madison, Madison, 53706, USA*

\* Corresponding author.

*E-mail address: sikai.chen@wisc.edu, nicole.chen@kaist.ac.kr*

**Abstract**

Semantic scene completion (SSC) plays a pivotal role in achieving comprehensive perception for autonomous driving systems. However, existing methods often neglect the high deployment costs of SSC in real-world applications, and traditional architectures such as 3D Convolutional Neural Networks (3D CNNs) and self-attention mechanisms struggle to efficiently capture long-range dependencies within 3D voxel grids, limiting their effectiveness. To address these challenges, we propose MetaSSC, a novel meta-learning-based framework for SSC that leverages deformable convolution, large-kernel attention, and the Mamba (D-LKA-M) model. Our approach begins with a voxel-based semantic segmentation (SS) pretraining task, designed to explore the semantics and geometry of incomplete regions while acquiring transferable meta-knowledge. Using simulated cooperative perception datasets, we supervise the training of a single vehicle's perception using the aggregated sensor data from multiple nearby connected autonomous vehicles (CAVs), generating richer and more comprehensive labels. This meta-knowledge is then adapted to the target domain through a dual-phase training strategy—without adding extra model parameters—ensuring efficient deployment. To further enhance the model's ability to capture long-sequence relationships in 3D voxel grids, we integrate Mamba blocks with deformable convolution and large-kernel attention into the backbone network. Extensive experiments show that MetaSSC achieves state-of-the-art performance, surpassing competing models by a significant margin, while also reducing deployment costs.

**Keywords**

3D semantic scene completion, Autonomous driving, Meta-learning, Deformable large-kernel attention, State space model

**Introduction**

Autonomous driving (AD) has seen rapid advancements, with research exploring different parts from perception (Li et al., 2023; Mei et al., 2024), planning (Sheng et al., 2024a, 2024b), localization (Huang et al., 2024), and control (Huang, Sheng, & Chen, 2024; Huang, Sheng, Ma, et al., 2024; Sheng et al., 2024). Among these, perception plays a foundational role, as it enables autonomous vehicles to sense, interpret, and understand their environment in real time. However, the suddenness and variability of dynamic traffic participants, combined with the extensive range and distance of static objects, present substantial challenges for autonomous vehicles in perceiving complex driving scenes. In many urban areas, these challenges are amplified by unpredictable behaviors from other drivers, cyclists, and pedestrians. Additionally, complex intersections, varying traffic signals, road geometry, and weather-induced changes further complicate perception. Autonomous vehicles must continuously analyze their surroundings and respond appropriately within milliseconds, making real-time perception essential for safe and effective operation. These challenges hinder the advancement of higher levels of autonomy, where vehicles need to operate with minimal human intervention.

Among the many approaches aimed at improving perception, Scene Semantic Completion (SSC) stands out as a technique that simultaneously reasons about both the geometry and semantics of a driving scene. As shown in Fig. 1, unlike traditional perception tasks that rely on individual object detection and tracking, SSC provides a more holistic understanding of the environment by filling in missing information from partial or occluded sensor inputs. This capability is especially critical when sensors like LiDAR or cameras are obstructed by other vehicles or environmental elements, as it helps ensure robust situational awareness. Recent progress in 3D SSC has demonstrated promising results across multiple domains (Cao & Behnke, 2024; Cheng et al.; Rist et al., 2021; Yang et al.) , laying the foundation for further research in this area.

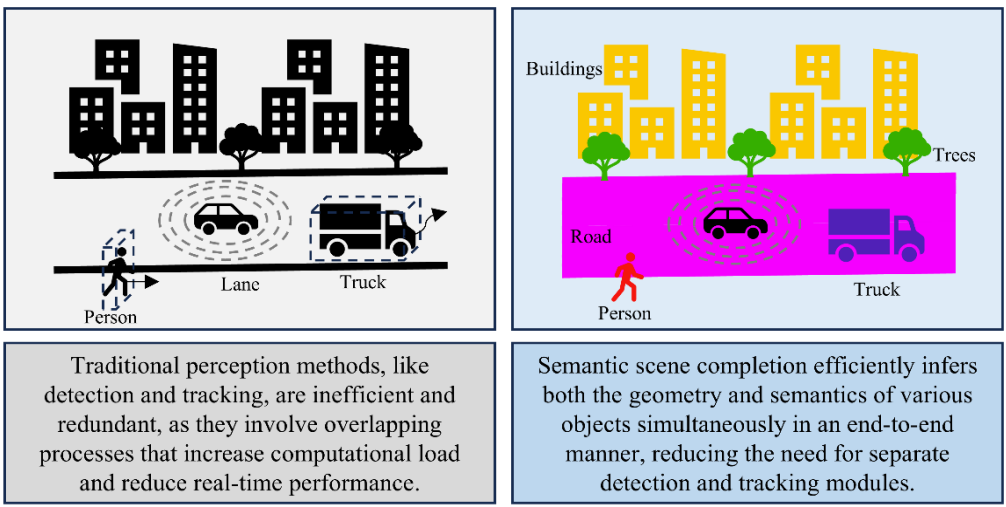

**Fig. 1.** The advantages of semantic scene completion over traditional perception tasks

However, collecting and labeling large-scale real-world datasets is a bottleneck in autonomous driving development. It is an expensive and labor-intensive process, often requiring hours of manual annotation for each scene. In addition, the diversity of real-world traffic situations means that capturing a sufficiently representative dataset is challenging. Some rare but critical scenarios, such as pedestrian collisions or vehicle malfunctions, may be impossible to gather at scale through real-world data collection alone. To address these challenges, researchers have increasingly turned to high-fidelity simulators such as CARLA (Dosovitskiy et al., 2017) and LGSVL (Rong et al., 2020) to generate synthetic data for model training. These simulators can replicate a wide range of scenarios, including adverse weather conditions, dense urban traffic, and rural environments, which helps bridge the gap between simulated and real-world driving scenarios. Moreover, simulation allows for faster data generation and supports experimentation with scenarios that are dangerous or rare in real life, such as near-collisions. However, transferring models trained on synthetic data to real-world tasks remains a non-trivial challenge due to the domain gap between simulated and real environments.

Additionally, current SSC solutions typically rely on 3D Convolutional Neural Networks (3D CNNs) to encode input data, such as point clouds or RGB-D images, which contain rich spatial information. However, 3D CNNs face challenges in capturing fine-grained scene representations or modeling long-sequence relationships between 3D blocks, both of which are essential for SSC. The lack of temporal modeling limits their ability to track dynamic changes in the environment, such as a vehicle's trajectory over time.

In response to these limitations, our study aims to address two critical gaps: (1) the need to efficiently leverage simulated data for fast deployment in real-world scenarios, and (2) the development of a novel backbone capable of capturing long-sequence dependencies and high-resolution spatial information. The proposed method seeks to bridge the gap between simulation and real-world applications while enhancing the efficiency of SSC in diverse driving environments.

The primary contributions of this study are summarized as follows:

(1) Dual-Phase Training with Meta-Learning

We adopt a dual-phase training strategy by pretraining the model on source domains (datasets generated from simulators) and fine-tuning it on the target domain (real-world datasets) using Model-Agnostic Meta-Learning (MAML) (Finn et al., 2017). This approach accelerates adaptation to real-world environments by quickly learning domain-specific features during fine-tuning. By generalizing across multiple domains, MAML reduces overfitting and improves the model's robustness in novel situations.

(2) A Novel Backbone for Long-Sequence Modeling

We introduce a new backbone architecture that integrates Mamba (Gu & Dao, 2023), a selective State Space Model (SSM), with Deformable Convolution (Dai et al., 2017) and Large-Kernel Attention (D-LKA) (Guo et al., 2023). Mamba provides a structured mechanism for handling sequential data over time, ensuring that long-range dependencies within 3D voxel grids are effectively captured. Deformable convolutions allow the model to dynamically adjust the receptive fields, enhancing its ability to detect objects at varying scales. Meanwhile, D-LKA enhances the network's attention mechanism, focusing on critical areas of the scene, which improves both spatial awareness and decision-making.

The remainder of this paper is organized as follows. Section 2 reviews related work on SSC, perception for autonomous driving, and transfer learning from simulated to real-world environments. Section 3 details the proposed methodology, including the backbone architecture and training strategies. Section 4 presents experimental results, benchmarking our approach against state-of-the-art methods and demonstrating its performance improvements. Section 5 concludes the paper, summarizing key findings and outlining future research directions.

**Related Work**

*2.1. 3D semantic scene completion for autonomous driving*

SSC infers both the geometry and semantics of large-scale outdoor environments from incomplete sensor inputs (Song et al., 2017) simultaneously. It offers a full understanding of driving scenes and predicts missing elements, which is critical for autonomous driving.

Roldão et al. (Roldao et al., 2020) proposed LMSCNet, a multiscale network that combines a 2D U-Net backbone with a 3D segmentation head. This design reduces the computational burden of full 3D convolutions while maintaining competitive performance. Similarly, Yan et al. (Yan et al., 2021) introduced a multi-task learning framework where Semantic Segmentation (SS) and SSC are trained jointly. By sharing features between both tasks, the model improves both geometry and semantic predictions.

Further studies leverage 2D-to-3D feature projections supervised by LiDAR point clouds (Cao & De Charette, 2022; Li, Yu, et al., 2023; Zhang et al., 2023). These approaches utilize monocular RGB cameras compared to LiDAR, which can lower the cost of deployment. However, transforming 2D RGB inputs into pseudo-voxelized point clouds presents challenges. False features can be introduced during this kind of pixel-to-point conversion in unoccupied areas of the 3D space, degrading model performance.

To address these limitations, recent research centers on improving pixel-to-point transformations and refining feature fusion techniques. Some methods incorporate depth estimation into RGB inputs, while others use attention mechanisms to selectively enhance relevant features. Despite these advancements, balancing accuracy and efficiency remains a challenge, especially for autonomous driving.

## 2.2. Sim2real knowledge transfer

Simulated datasets provide scalable and diverse environments for testing algorithms, helping overcome the limitations of real-world data collection (Qu et al., 2024; Qu et al., 2023). For example, Xu et al. (Xu et al., 2022) introduced the OPV2V dataset, which facilitates the evaluation of multiple object detection methods and fusion strategies for Vehicle-to-Vehicle (V2V) perception in CARLA (Dosovitskiy et al., 2017). Similarly, Li et al. (Li et al., 2022) proposed the V2X-SIM datase on Vehicle-to-Everything (V2X) perception for autonomous driving. These datasets have become the benchmarks and essential tools for advancing cooperative perception research in autonomous driving.

However, there is always a significant "reality gap" existing between simulated environments and real-world conditions. Models trained on data collected from simulated environments might perform well on simulated environments, but often struggle to perform reliably in real-world settings due to domain gap. To bridge this gap, researchers have explored meta-learning approaches (Hu et al., 2023), where MAML (Finn et al., 2017) has gained attention for its ability to leverage prior experiences, enabling rapid adaptation to new tasks by fine-tuning with minimal data of target domain. This makes MAML particularly well-suited for sim-to-real transfer, where models need to generalize quickly to new conditions.

Building on the success of sim-to-real methods (Kong et al., 2023; Tanwani, 2021; Zhao et al., 2021) and meta-learning frameworks (Lu et al., 2022), we explore the feasibility of identifying a pretraining task closely aligned with SSC while preserving the focus on incomplete data. Inspired by cooperative perception (Song et al., 2024; Zhang et al., 2024), we design the pretraining task as a cooperative voxel semantic segmentation task, where each ego vehicle uses its sensor data to infer the full scene, and the supervision is provided by the combined sensor data from all Connected Autonomous Vehicles (CAVs) in the simulation. This cooperative framework enhances scene understanding by utilizing meta-knowledge, offering a promising strategy for reducing the deployment cost of perception models in real-world settings.

*2.3. Deformable large kernel attention*

There are two primary methods for learning the correlations between different voxels in the SSC task (Guo et al., 2023) . The first approach utilizes 3D convolutions with large kernels and stacking multiple layers, enabling the model to capture long-range dependencies across the 3D space. However, the computational cost increases exponentially with the number of layers, and the large number of parameters requires more memory and training time. These limitations make it impractical for real-time applications, especially in autonomous driving scenarios, where efficiency is crucial.

The second approach utilizes self-attention mechanisms, selectively attending to relevant features. self-attention provides flexibility in modeling relationships between distant voxels. However, self-attention tends to overlook the inherent 3D structure of the scene, treating the input data more like flattened sequences rather than structured spatial information. Besides, self-attention does not dynamically adapt to changes in the channel dimension, limiting its ability to represent complex transformations in driving environments. These limitations, combined with the computational overhead of attention-based models, present challenges for deploying them in resource-constrained systems.

To address these issues, researchers have explored deformable convolutions (Dai et al., 2017; G. Wang et al., 2024), which introduce additional offsets that allow the network to adaptively resample spatial features. This approach enhances the model's ability to handle geometric variations by focusing on the most relevant regions of the input, improving its robustness in complex scenarios. Deformable convolutions have proven effective in tasks where objects undergo non-rigid transformations, making them particularly useful for autonomous driving.

Building on previous work (Azad et al., 2024), we applied Deformable Large Kernel Attention (D-LKA) to SSC tasks. This hybrid approach aims to combine the strengths of both 3D convolutions and attention mechanisms, enabling the model to capture long-range dependencies while maintaining awareness of the 3D structure (Jiang et al., 2023). The deformable component ensures adaptability to geometric variations, while the large-kernel attention improves the focus on critical regions of the scene. This combination enhances the model's ability to represent complex driving environments effectively, supporting better SSC performance.

### *2.4. Mamba on 3D semantic scene completion*

Due to its computational efficiency, Mamba (Gu & Dao, 2023) has emerged as a promising successor to the Transformer (Vaswani et al., 2017) and has recently gained significant attention (X. Wang et al., 2024). Mamba's streamlined architecture reduces the computational overhead typically associated with Transformers, making it well-suited for applications that require fast inference. It adopts a lightweight design, which replaces the multi-head self-attention mechanism with simpler linear transformations, while still capturing essential relationships between input elements. As a result, Mamba has proven to be an effective model for tasks that involve sequential data, where traditional Transformers can be too resource-intensive.

For instance, Zhu et al. (Zhu et al., 2024) developed a generic vision backbone based on Mamba to model the relationships between image patches, demonstrating the potential of Mamba for computer vision tasks. By efficiently encoding the relationships between image regions, Mamba offers a practical alternative to Transformer-based models in visual processing. Furthermore, Mamba could be even more effective in 3D modeling tasks, where the sequence of 3D blocks is substantially longer and more complex than 2D image patches. This insight has encouraged researchers to explore new ways of extending Mamba's capabilities beyond 2D applications.

One such effort is OccMamba (Li et al., 2024), which introduces a 3D-to-1D re-ordering operation to adapt Mamba for 3D modeling. Specifically, a height-prioritized 2D Hilbert expansion is used to transform 3D voxel grids into linear sequences, enabling the application of Mamba's linear modeling capabilities directly to 3D data. Moreover, several studies (Fei et al., 2024; Lieber et al., 2024; Z. Wang et al., 2024; Xu et al., 2024) have explored combining Transformer architectures with Mamba, suggesting that the two models complement each other. While Mamba excels in efficiency and scalability, Transformers provide robust capacity for capturing global dependencies. Integrating the strengths of both architectures offers new opportunities for enhanced performance among multiple tasks, particularly in 3D modeling.

**Methodology**

Research (Dai et al., 2018; Yan et al., 2021) has shown that combining SS and SSC within a multi-task learning framework enhances the performance of both tasks, where SS provides detailed semantic features that complement the geometric understanding captured by SSC, allowing both modules to benefit from shared feature extraction. Also, some approaches increase the density of semantic labels by using historical LiDAR scans as auxiliary supervision. While these methods improve the model's capacity to capture fine-grained semantics, the reliance on historical scans adds computational overhead, making these solutions challenging to deploy in real-time autonomous driving scenarios.

Our approach differs by treating SS as a pretraining task to learn meta-knowledge for SSC. The pretraining step helps the model generalize better across different domains, preparing it to handle real-world complexities such as occlusion and sensor noise. To further enhance supervision, we aggregate semantic information from nearby CAVs to provide denser labels that extend across greater distances. This aggregated semantic information from multiple vehicles addresses the limitations of individual sensors, which are often constrained by sparse data and occlusion. It allows the model to reason more effectively about incomplete areas, resulting in a more comprehensive scene understanding.

This strategy not only improves the performance of SSC but also reduces training costs. By pretraining the model with simulated data and leveraging CAV-based information in real-world scenarios, our approach minimizes the need for large annotated datasets, accelerating deployment. We adopt a dual-phase training strategy based on MAML (Finn et al., 2017), which consists of a meta-training phase and an adaptation phase. In the meta-training phase, the model learns transferable knowledge from SS using simulation data, building a foundation of generalized features. During the adaptation phase, the model is fine-tuned on real-world data for SSC, allowing it to refine its understanding of specific scenarios it encounters (Section 3.1).

Fig. 2 visualizes this concept, illustrating how both $X$ and $Y_{ego}$ are sparser compared to $Y_{cavs}$, which offer higher occupancy and provide better exploration of incomplete positions. The sparser inputs from ego vehicles highlight the limitations of single-vehicle perception, whereas the dense labels from surrounding CAVs provide more comprehensive supervision, filling in missing details that the ego vehicle alone could not capture. In practice, we leverage $Y_{cavs}$ as supervision for SSC to overcome the limitations of sparse input data.

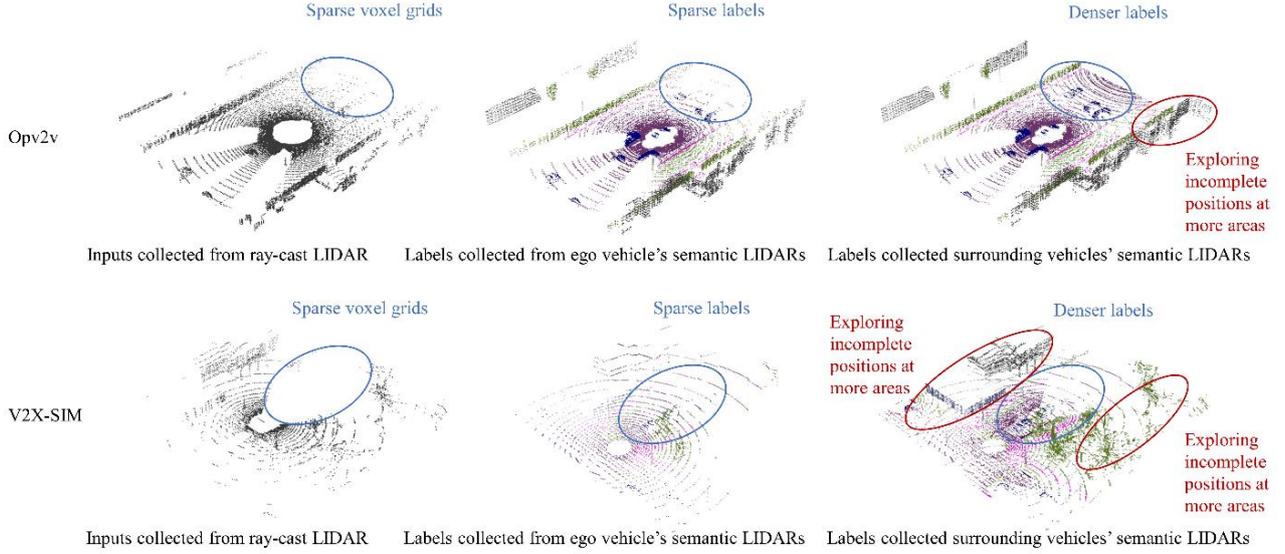

**Fig. 2.** Aggregating semantics information from surrounding CAVs to provide richer context information and denser labels.

This dual-phase strategy ensures robust performance improvements and allows for efficient training, benefiting both simulation and real-world applications. Furthermore, inspired by recent developments (Fei et al., 2024; Lieber et al., 2024; Z. Wang et al., 2024; Xu et al., 2024), we integrate deformable convolution, large-kernel attention, and Mamba to tackle the challenge of long-sequence modeling in 3D blocks.

*3.1. Problem formulation*

We define the problem of 3D SSC as follows: Given a sparse 3D voxel grid $X \in R^{H \times W \times D}$, where $H$, $W$, and $D$ denote the height, width, and depth of the driving scene, respectively. Each voxel $x_{i,j,k}$ in $X$ can be either 0 or 1, indicating the occupancy of objects, where $i, j$ and $k$ are the voxel indexes. The objective of 3D SSC is to learn a model $f(*)$ that assigns a semantic label to each $x_{i,j,k}$ in $X$, resulting in $C = f(X)$, where $c_{i,j,k}$ is the label at the corresponding position. These labels belong to the set $[c_0, c_1, \ldots, c_N]$, with $N$ being the number of semantic classes and $c_0$ representing a free label.

*3.2. Dual-phase training strategy*

Based on MAML (Finn et al., 2017), the workflow of our proposed method, MetaSSC, is illustrated in Fig. 2 and consists of two main phases: meta-pretraining and adaptation. These phases enable the MetaSSC model to transfer knowledge from simulated environments to real-world driving scenarios, improving performance on 3D SSC tasks.

The meta-pretraining phase (Fig. 2-Part A) aims to prepare the model for generalization across diverse tasks by learning from simulated data. The source datasets, OPV2V (Xu et al., 2022) and V2XSIM (Li et al., 2022), provide a range of V2V and V2X scenarios that help the model develop robust features for dynamic environments. Tasks are sampled from these datasets, each comprising a support set and a query set. The support set is used to optimize task-specific parameters in the inner loop, while the query set evaluates the generalization performance of the model in the outer loop.

The meta-learner initializes the MetaSSC backbone with a set of parameters θ, which are assigned to each task. Given n tasks $\mathcal{T}_1 = S_1 \cup Q_1, \mathcal{T}_2 = S_2 \cup Q_2, \ldots, \mathcal{T}_n = S_n \cup Q_n$ from source datasets $\mathcal{D}_S$, for each task $\mathcal{T}_i$, the support set $S_i$ is used in the inner loop, where multiple k-step gradient descents are performed. These small-step updates allow the model to quickly adapt to task-specific features, improving its ability to handle complex scenarios. During each step, data augmentation (Aug) is applied to enhance the robustness and generalization of the learned features.

After the inner loop, the query set $Q_i$ evaluates the model's generalization, and the outer loop updates are applied. Algorithm 1 summarizes this process. As shown, for each task, the meta-learner assigns the model parameters as

$$\theta_i \leftarrow \theta, \tag{1}$$

where $\theta_i$ is initialized to the meta-learner's parameters. For $k$-steps, the parameters are updated using the inner-loop gradient descent:

$$\theta_i \leftarrow \theta_i - \alpha \nabla_{\theta_i} L_i^{\text{src}}(f_{\theta_i}, S_i), \tag{2}$$

where $\alpha$ is the step size for the inner-loop update, and $L_i^{\text{src}}$ is the loss function for task $\mathcal{T}_i$ on the source dataset.

---

**Algorithm 1 Meta pretraining**

1: **Require:** $\alpha$: The step sizes for the gradient update of inner loop. $\beta$: The step size for the gradient update of outer loop.
2: **Require:** The source dataset $\mathcal{D}_S$.
3: **Require:** $\mathcal{L}_i^{src}$: The loss function of task $\mathcal{T}_i$ on source datasets.
4: Randomly initialize $\theta$;
5: **while** not done **do**
6:    Sample $n$ source tasks $\mathcal{T}_1 = S_1 \cup Q_1, \mathcal{T}_2 = S_2 \cup Q_2, \ldots, \mathcal{T}_n = S_n \cup Q_n$ from $\mathcal{D}_S$;
7:    **For all $\mathcal{T}_i$ do**
8:       Assign model parameters by $\theta_i \leftarrow \theta$
9:       **For $k$ steps do**
10:          $\theta_i \coloneqq \theta_i - \alpha \nabla_{\theta_i} \mathcal{L}_i^{src}(f_{\theta_i}, S_i)$
11:       **end for**
12:       Use $\theta_i$ as the initial set of model parameters for the next task
13:       Validate on query set to get $\mathcal{L}_i^{src}(f_{\theta_i}, Q_i)$
14:    **end for**
15:    Update meta-learner by $\theta^* = \theta_n - \beta \nabla_{\theta_i} \sum_{i=1}^n \mathcal{L}_i^{src}(f_{\theta_i}, Q_i)$
16: **end while**

---

After the inner loop completes, the query set $Q_i$ evaluates the updated parameters, and the outer-loop loss $L_i^{\text{src}}(f_{\theta_i}, Q_i)$ is computed. These losses from all tasks are accumulated, and the final meta-update is performed as:

$$\theta^* \leftarrow \theta_n - \beta \nabla_{\theta_i} \sum_{i=1}^{n} L_i^{\text{src}}(f_{\theta_i}, Q_i) \tag{3}$$

where $\beta$ is the step size for the outer-loop gradient update. The meta-pretraining phase actually balances the updates across all tasks, resulting in model parameters that are moderately fitted for all tasks. This meta-updated model becomes the initialization for the next training epoch, allowing the model to improve across tasks continuously.

The workflow in Fig. 3 Part A shows how multiple tasks (Task 1 to Task n) are processed in parallel, with inner-loop updates for each task followed by outer-loop updates. This iterative process ensures that the model generalizes well across different tasks, preparing it for real-world adaptation.

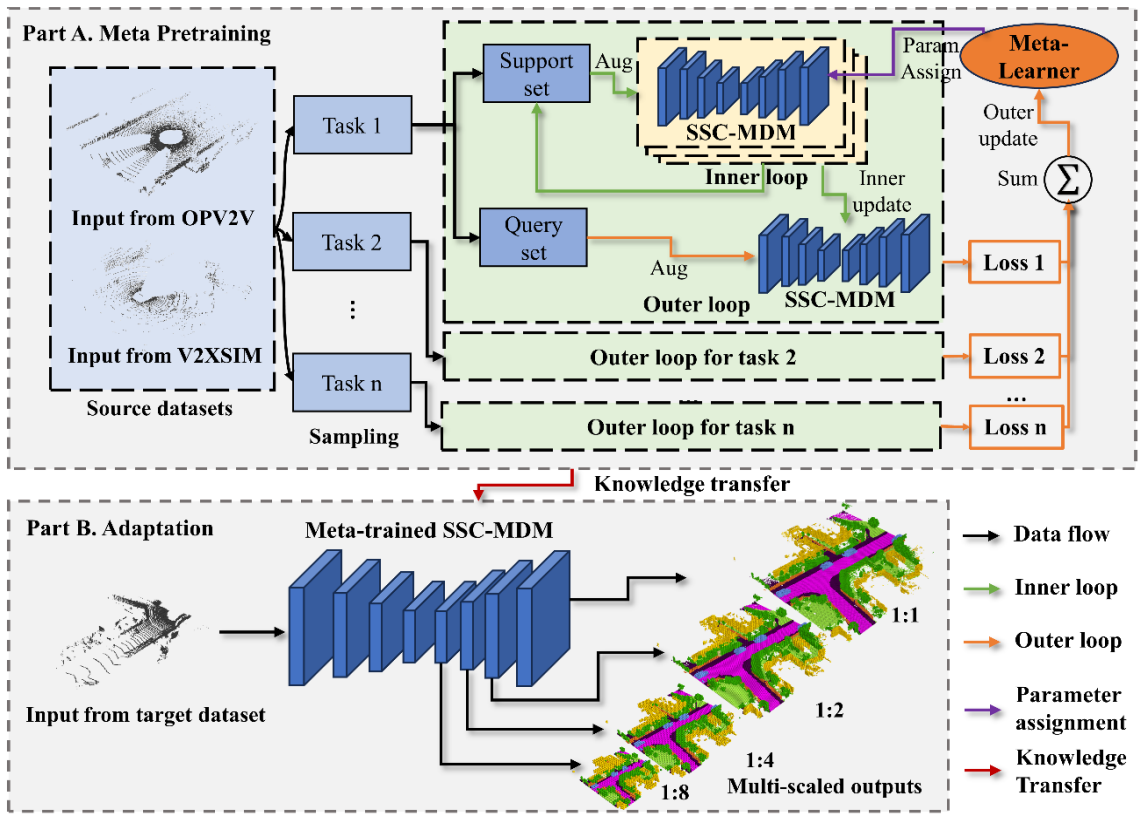

**Fig. 3.** The workflow of MetaSSC.

In the adaptation phase (Fig. 2-Part B), the meta-trained MetaSSC model is adapted to the target real-world dataset, SemanticKITTI (Behley et al., 2019). This phase fine-tunes the meta-learned parameters to align with real-world conditions, addressing challenges such as sensor noise, occlusion, and environmental variability. Multi-scale completion, as described in (Roldao et al., 2020), allows the model to generate outputs at various resolutions (1:1, 1:2, 1:4, and 1:8), enabling it to capture both fine details and large-scale features of the driving environment.

The multi-scaled outputs are crucial for balancing local precision and global scene understanding. For example, smaller objects like pedestrians are detected at finer scales, while larger objects like roads and buildings are identified at coarser resolutions. This hierarchical output structure ensures that the model provides accurate and comprehensive scene completion, even in challenging real-world scenarios.

The adaptation phase leverages the meta-learned parameters as a strong starting point, minimizing the need for extensive retraining. This efficient transfer learning framework accelerates the deployment of the MetaSSC model in real-world settings, ensuring high performance with minimal computational overhead.

Finally, we also summarize the meta-pretraining and adaptation phase in Algorithm 1 and Algorithm 2.

| Algorithm 2 Adaptation |
| --- |
| 1: **Require:** $\lambda$: The step sizes for the gradient update of adaptation. |
| 2: **Require:** The target dataset $\mathcal{D}_T$. |
| 3: **Require:** $\mathcal{L}_i^{tgt}$: The loss function of task $\mathcal{T}_i$ on the target dataset. |
| 4: Assign the model parameters $\theta \leftarrow \theta^*$ using pre-trained parameters from meta-learner; |
| 5: **for all** $\{X_i, Y_i\} \in \mathcal{D}_T$ **do** |
| 6:     Compute updated model parameter for data imputation: $\theta \coloneqq \theta - \lambda \nabla_\theta \mathcal{L}^{tgt}(f_\theta, X_i)$ |
| 7: **end for** |

### 3.3. D-LKA-M architecture

The architecture of D-LKA-M, as shown in Fig. 4, is derived from the D-LKA network (Azad et al., 2024) with the integration of Mamba blocks (Gu & Dao, 2023) to effectively handle the long-sequence modeling of 3D blocks. This design follows a hierarchical structure similar to LMSCNet (Roldao et al., 2020)[10] , resembling a U-Net (Ronneberger et al., 2015) architecture. The hierarchical structure enables multi-scale processing, allowing the model to capture both fine-grained details and broader contextual information from 3D scenes.

The model processes input data through a series of 3D modules, with down-sampling and up-sampling operations at different stages of the network. Each down-sampling layer reduces the spatial dimensions, compressing the input while retaining essential information, and each up-sampling layer reconstructs higher-resolution outputs. This structure makes it capable of outputting results at various reduced resolutions. It is particularly useful for SSC tasks, as it provides predictions at multiple scales, improving the accuracy in SSC.

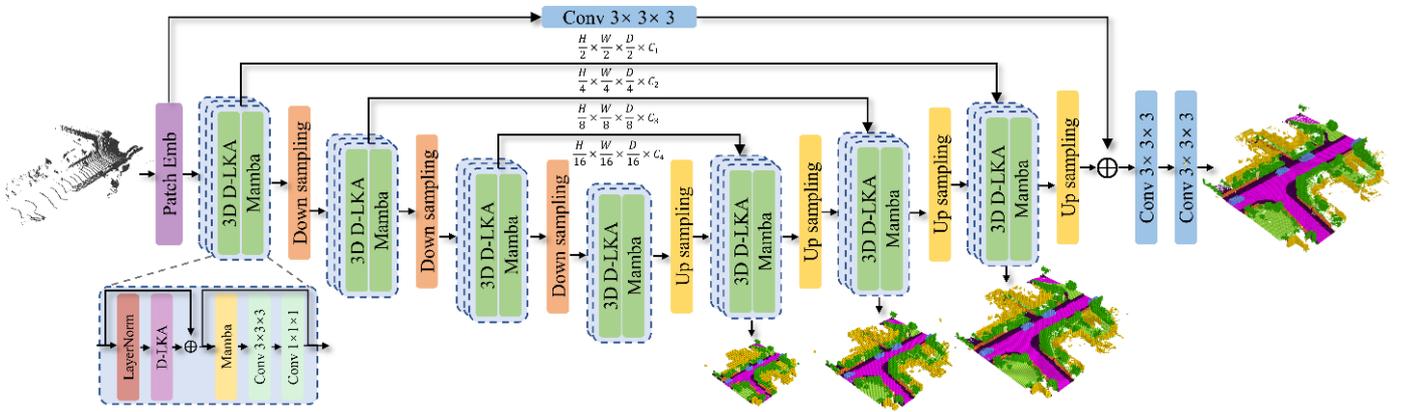

**Fig. 4.** The architecture of D-LKA-M.

Patch embedding modules are used at the input stage to divide the raw 3D data into manageable parts. The Mamba blocks, embedded within the D-LKA modules, enhance the network's ability to model long-range dependencies across 3D voxel grids, which is essential for understanding complex driving environments. This integration ensures that the model balances both computational efficiency and accuracy, making it suitable for real-time applications.

*3.3.1 Deformable convolution*

Deformable convolution (Dai et al., 2017) introduces an offset field to adjust the convolution kernel adaptively, which is especially crucial in autonomous driving, where objects like pedestrians, vehicles, and obstacles often do not conform to rigid shapes or positions. It can be summarized as follows (Azad et al., 2024):

$$x_1 = DAttn(LN(x_{in})) + x_{in} \qquad (4)$$

where $DAttn(*)$ denotes the deformable attention mechanism, $LN(*)$ denotes the layer normalization.

In deformable attention, for any position $p_0$ in the input feature map $x$, the learned offset $\Delta p_n$ is added to the receptive field, defined as $p_0 + p_n$. Here, $p_n$ enumerates the positions in a regular voxel grid. The deformable convolution output at position $p_0$ is given by:

$$y(p_0) = \sum_{p_n \in \mathcal{R}} w(p_n) \cdot x(p_0 + p_n + \Delta p_n) \qquad (5)$$

Since the offset $p_n$ is often fractional, interpolation is required to compute the feature values at non-integral positions. The interpolated value at position $p$ is computed as:

$$x(p) = \sum_q G(q, p) \cdot x(q) \qquad (6)$$

In summary, deformable convolution provides significant advantages for autonomous driving by improving the model's capacity to understand complex scenes. This is essential for building safer and more reliable autonomous driving systems capable in real-world settings.

*3.3.2 Large-kernel attention*

Inspired by (Azad et al., 2024), we use deformable convolution and large kernel attention as a basic module (D-LKA) in our model. The deformable convolution provides adaptive receptive fields to handle irregular object shapes, while LKA ensures efficient processing of both local details and global context. Together, D-LKA enhances the model's ability to accurately capture complex spatial relationships within the 3D voxel grids.

Specifically, the large-kernel attention (LKA) (Guo et al., 2023) is decomposed into a $\left\lceil \frac{K}{d} \right\rceil \times \left\lceil \frac{K}{d} \right\rceil$ depth-wise dilation convolution with dilation $d$, a $(2d-1) \times (2d-1)$ depth-wise convolution, and a $1 \times 1$ channel convolution. The mathematical formulation of LKA can be expressed as:

$$A = Conv_{1\times 1}\left(DWDConv(DWConv(x))\right) \tag{7}$$

where $x$ is the input feature, DWConv denotes the depth-wise convolution, DWDConv represents the depth-wise dilation convolution, and $Conv_{1\times 1}$ is the channel convolution. The final output of LKA is obtained through an element-wise product $x_{out} = A \otimes x$ between the attention weights $A$ and the input feature $x$, where A represents the attention weights and $\odot$ denotes the element-wise product.

Combining deformable convolution and large kernel attention plays a crucial role in enabling the model to perform effectively in autonomous driving scenarios, where both local details and large-scale context are essential.

*3.3.3 Mamba*

Unlike Vision Mamba (Zhu et al., 2024), which applies Mamba blocks within a purely 2D feature extraction pipeline, our approach directly processes the features learned from the D-LKA block with the Mamba block (Gu & Dao, 2023) to enhance long-sequence modeling for 3D voxel grids. The mathematical formulation of this process is expressed as $x_2 = Mamba(x_1)$, where $x_1$ represents the input features extracted from the D-LKA block, and $Mamba(*)$ denotes the Mamba block.

Once the features are processed by the Mamba block, they are further refined through a feed-forward network (FFN) and a convolution layer. The final output $x_{out}$ is computed as:

$$x_{out} = Conv(FFN(x_1)) + x_2 \tag{8}$$

where $Conv(*)$ represents the convolution layer, and $FFN(*)$ is a feed-forward network.

In summary, the integration of D-LKA and Mamba blocks allows our model to perform both local and long-sequence modeling efficiently, and also ensures a balance between local details and global context for accurate decision-making.

**Experiments**

To identify a pretraining task akin to SSC while preserving its focus on exploring incomplete positions, we utilized OPV2V (Xu et al., 2022) and V2X-SIM (Li et al., 2022) as source domains to meta-train the D-LKA-M model. The aim was to create a pretraining process that closely mimics SSC's challenges, such as the incomplete perception of the driving environment, encouraging the model to develop robust completion strategies for real-world scenarios. Since OPV2V does not contain semantic labels in its original datasets, we positioned five semantic LIDARs at the sites of the original cameras and ray-cast LIDAR, replaying the simulations to capture the semantics of driving scenes within the cameras' Field of View (FoV). This adjustment ensures that the necessary semantic information is available for training, replicating the conditions needed for scene completion. For V2X-SIM, we used the originally provided semantic data without modifications. However, both OPV2V and V2X-SIM only capture the ego vehicle's surrounding semantics, neglecting the semantics of other moving objects. To address this, we associated ground-truth labels from all CAVs but continued using the ego vehicle's ray-cast LIDAR as the primary input. This strategy ensures that the model learns to infer a more complete scene understanding from limited input, aligning well with the objective of SSC to explore and predict incomplete positions.

We conducted experiments on SemanticKITTI (Behley et al., 2019), a real-world outdoor dataset with urban scenes, providing 3D voxel grids from semantically labeled scans of the HDL-64E rotating LiDAR (Geiger et al., 2012). These scans represent real-world driving environments and test the model's ability to transfer knowledge gained from simulated data to real scenarios. The voxel grids are highly sparse, with a shape of 256×256×32 and a voxel size of 0.2m, presenting challenges in accurately completing scenes from incomplete inputs. This sparsity requires the model to perform efficiently with limited data, mimicking real-world autonomous driving conditions. We used the original SemanticKITTI settings (Behley et al., 2019) to split the data into training, validation, and testing sets, ensuring consistency with previous studies.

We employed the Adam optimizer (Kingma, 2014) with an initial learning rate of 0.001 for most experiments; for the meta-training stage, we used Stochastic Gradient Descent (SGD). All experiments were performed on an Ubuntu operating system with Python 3.8, and the code was implemented using PyTorch. All models were trained on a Dell Precision T3650 workstation equipped with a 24 GB Nvidia GeForce RTX 3090 Graphics Card and an Intel I7-10700K CPU.

*4.1 Experiment 1: comparison with baseline models*

The baseline models are introduced as follows.

**(1) SSCNet**

SSCNet (Song et al., 2017) is an end-to-end 3D convolutional network designed to take a single depth image as input and perform scene completion by generating semantic voxel predictions.

**(2) TS3D**

TS3D (Garbade et al., 2019) follows a two-stream design, which allows it to separate the processing of geometry and semantics, resulting in detailed scene predictions. However, its performance heavily depends on additional RGB inputs, which may not always be available in real-world applications.

**(3) LMSCNet**

LMSCNet (Roldao et al., 2020) employs a 2D U-Net backbone with multi-scaled connections to extract features from the input data. It predicts both occupancy and semantic labels simultaneously using 3D segmentation heads, achieving improved scene understanding. The multi-scaled connections provide the model with the ability to capture features at different levels of granularity.

**(4) PointPainting**

Vora et al. introduce PointPainting (Vora et al., 2020), a sensor fusion approach that enhances lidar point clouds by integrating semantic class scores obtained from an image-based segmentation network, enabling the processed point cloud to be utilized by any lidar-only detection model.

**(5) JS3CNet**

JS3CNet (Yan et al., 2021) is a multi-task framework designed to integrate SS and SSC. It employs a Point-Voxel Interaction module to facilitate the fusion of point-level and voxel-level knowledge between the two tasks. The SSC module further leverages contextual shape priors, which help the network better understand object structures and improve performance in scene completion.

**(6) UDNet**

UDNet (Zou et al., 2021) utilizes multi-scaled context information to perform SSC at multiple resolutions, enhancing its ability to predict complex scenes with high accuracy. By working across different resolutions, UDNet is capable of capturing both fine details and large-scale features within the driving environment.

**(7) MonoScene**

MonoScene (A.-Q. Cao & De Charette, 2022) introduces a 3D SSC framework that reconstructs dense geometry and semantic information of a scene from a single monocular RGB image. The method employs sequential 2D and 3D UNets connected through a novel 2D-to-3D feature projection inspired by optical principles and incorporates a 3D context relation prior to ensure spatial and semantic consistency.

**(8) TPVFomer**

This paper (Huang et al., 2023) proposes a tri-perspective view (TPV) representation for 3D scene perception, combining bird's-eye-view (BEV) with two additional perpendicular planes to capture fine-grained 3D structures. To map image features to the TPV space, it introduces a transformer-based TPV encoder, which uses attention mechanisms to aggregate features across planes, enabling effective semantic occupancy prediction for all voxels with sparse supervision.

**(9) OccFormer**

OccFormer (Zhang et al., 2023) introduces a dual-path transformer network for 3D semantic occupancy prediction, efficiently processing 3D voxel features generated from camera inputs. The method decomposes 3D computation into local and global transformer pathways along the horizontal plane for dynamic and long-range feature encoding. Additionally, the occupancy decoder adapts Mask2Former with preserve-pooling and class-guided sampling to address sparsity and class imbalance issues.

The results of our proposed MetaSSC model are compared against these baseline models on the SemanticKITTI benchmark (Behley et al., 2019), as summarized in Table 1.

The experimental results demonstrate that the proposed MetaSSC outperforms existing methods in several key metrics, particularly achieving higher overall performance with an mIoU of 20.7 and IoU of 59.5, surpassing strong baselines such as UDNet, TPVFormer, and Monoscene. MetaSSC exhibits exceptional accuracy in common classes like Road and Building, showcasing its strong capability in capturing both the structural and semantic information of the scene. These results validate the effectiveness of the proposed framework, which combines meta-learning and large-kernel attention mechanisms and Mamba model.

However, MetaSSC's performance on the SSC task is slightly lower than that of JS3CNet (mIoU: 23.8). This can be attributed to JS3CNet's voxel-based architecture, which is specifically tailored for dense SSC tasks. In contrast, MetaSSC emphasizes computational efficiency and adaptability, which may result in some loss of fine-grained spatial details. Despite this trade-off, MetaSSC offers broader applicability and maintains a competitive performance in most metrics.

The model also encounters challenges in detecting small and rare objects, such as Bicycles, Motorcycles, and Poles. These difficulties are likely due to the sparse representation of such objects in the training data and the inherent sparsity of the 3D voxel space. Addressing these issues may require targeted improvements in data augmentation or loss functions designed specifically for rare and small-scale objects. Additionally, our method employs a two-stage meta-learning framework, consisting of meta-pretraining and adaptation phases. In the meta-pretraining stage, we use the OPV2V (Xu et al., 2022) and V2XSIM (Li et al., 2022) datasets, which do not include detailed representations of rare object categories. Consequently, the meta-pretraining phase may result in incomplete feature representations for these rare objects. This limitation in pretraining affects the adaptation phase, where the model may struggle to refine these representations for such categories, impacting the final performance.

The results also highlight the varying performance of models with different input modalities. For example, TS3D, which incorporates RGB inputs, achieves higher recall due to the additional contextual information from RGB data. In contrast, MetaSSC, relying solely on LiDAR-based inputs, demonstrates its robustness and suitability for real-world autonomous driving scenarios.

In summary, the proposed MetaSSC framework achieves state-of-the-art performance on IoU and mIoU, and excels in handling common and structurally significant objects. While it has limitations in specific areas such as small object detection and SSC compared to JS3CNet, its adaptability, computational efficiency, and overall performance make it a highly robust and effective solution for semantic scene completion in autonomous driving.

**Table 1.** Comparison with baseline models on the test dataset of SemanticKITTI (Behley et al., 2019) benchmark

| Approach | Year | Input | SS IoU | mIoU | Road | Sidewalk | Parking | Other-gr | Building | Car | Truck | Bicycle | Motorcycle | Other-veh | Vegetation | Trunk | Terrain | Person | Bicyclist | Motorcycli | Fence | Pole | Traffic- |
|---|---|---|---|---|---|---|---|---|---|---|---|---|---|---|---|---|---|---|---|---|---|---|---|
| SSCNet | 2017 | $x^{pc}$ | 51.4 | 16.8 | 60.2 | 32.6 | 26.2 | 6.9 | 35.5 | 27.2 | 1.4 | 0.0 | 0.1 | 5.7 | 34.2 | 18.7 | 29.6 | 0.0 | 0.0 | 0.0 | 20.8 | 14.0 | 5.8 |
| TS3D | 2019 | $x^{pc,rgb}$ | 29.8 | 9.5 | 28.0 | 17.0 | 15.7 | 4.9 | 23.2 | 10.7 | 2.4 | 0.0 | 0.0 | 0.2 | 24.7 | 12.5 | 18.3 | 0.0 | 0.1 | 0.0 | 13.2 | 7.0 | 3.5 |
| LMSCNet | 2020 | $x^{pc}$ | 55.2 | 16.6 | 63.2 | 33.0 | 26.2 | 2.5 | 38.0 | 29.6 | 0.2 | 0.0 | 0.0 | 0.0 | 40.7 | 18.0 | 31.0 | 0.0 | 0.0 | 0.0 | 18.3 | 14.4 | 0.2 |
| PointPainting | 2020 | $x^{pc,rgb}$ | 58.9 | 21.6 | 68.2 | 39.4 | 26.9 | 6.9 | 35.7 | 43.0 | 9.5 | 4.1 | 4.5 | 6.4 | 36.1 | 16.6 | 43.1 | 1.3 | 0.9 | 0.2 | 22.5 | 26.2 | 16.9 |
| JS3CNet | 2021 | $x^{pc}$ | 56.6 | 23.8 | 64.7 | 39.9 | 34.9 | 14.1 | 39.4 | 33.3 | 7.2 | 14.4 | 8.8 | 12.7 | 43.1 | 19.6 | 40.5 | 8.0 | 5.1 | 0.4 | 30.4 | 18.9 | 15.9 |
| UDNet | 2021 | $x^{pc}$ | 59.4 | 19.5 | 62.0 | 35.1 | 28.2 | 9.1 | 39.5 | 33.9 | 3.8 | 0.8 | 0.4 | 4.4 | 40.9 | 23.2 | 32.3 | 0.5 | 0.3 | 0.3 | 24.4 | 18.8 | 13.1 |
| Monoscene | 2022 | $x^{rgb}$ | 34.2 | 11.1 | 54.7 | 27.1 | 24.8 | 5.7 | 14.4 | 18.8 | 3.3 | 0.5 | 0.7 | 4.4 | 14.9 | 2.4 | 19.5 | 1.0 | 1.4 | 0.4 | 11.1 | 3.3 | 2.1 |
| TPVFormer | 2023 | $x^{rgb}$ | 34.3 | 11.3 | 55.1 | 27.2 | 27.4 | 6.5 | 14.8 | 19.2 | 3.7 | 1.0 | 0.5 | 2.3 | 13.9 | 2.6 | 20.4 | 1.1 | 2.4 | 0.3 | 11.0 | 2.9 | 1.5 |
| Occformer | 2023 | $x^{rgb}$ | 34.5 | 12.3 | 55.9 | 30.3 | 31.5 | 6.5 | 15.7 | 21.6 | 1.2 | 1.5 | 1.7 | 3.2 | 16.8 | 3.9 | 21.3 | 2.2 | 1.1 | 0.2 | 11.9 | 3.8 | 3.7 |
| IPVoxelNet | 2024 | $x^{pc}$ | 56.7 | 15.8 | 62.1 | 30.7 | 0.7 | 0.0 | 36.6 | 28.6 | 0.0 | 0.0 | 0.0 | 4.6 | 42.6 | 20.9 | 33.1 | 3.2 | 3.4 | 0.6 | 16.2 | 20.1 | 0.11 |
| MetaSSC (ours) | - | $x^{pc}$ | 59.5 | 20.7 | 66.5 | 39.4 | 33.5 | 8.7 | 41.3 | 33.7 | 3.9 | 0.4 | 0.5 | 5.7 | 43 | 22.7 | 34.3 | 1.4 | 0.6 | 0.6 | 28 | 18.4 | 10.3 |

We also add Precision and Recall as extra metrics (Table 2), re-evaluate the performance, and compare with selected baselines. We found MetaSSC model ranks 1st in Intersection over Union (IoU) for scene completion and 2nd in Precision. It also achieves 2nd place in mean Intersection over Union (mIoU) for SSC, indicating its superior performance in both scene completion and semantic scene completion tasks. However, the Recall of MetaSSC is lower than that of TS3D, which can be attributed to TS3D's use of additional RGB inputs. This difference highlights the trade-off between RGB-aided performance and purely LiDAR-based models like MetaSSC.

Regarding recall, we note that TS3D achieves a high recall score primarily due to its use of RGB image and point cloud inputs, which provide richer contextual information. In contrast, our method relies solely on LiDAR data, making it more efficient and practical for real-world applications. Despite this limitation in recall, our model outperforms TS3D in IoU, demonstrating its robustness in scene understanding and semantic completion.

**Table 2.** Comparison with baseline models on test dataset of SemanticKITTI (Behley et al., 2019) benchmark using Precision and Recall as extra metrics

| Approach | Precision | Recall | IoU | mIoU |
|---|---|---|---|---|
| SSCNet | 59.4 | 79.3 | 51.4 | 16.8 |
| LMSCNet | 80.5 | 63.8 | 55.2 | 16.6 |
| TS3D | 31.6 | 84.2 | 29.8 | 9.5 |
| JS3CNet | 71.5 | 73.5 | 56.6 | 23.8 |
| UDNet | 78.5 | 70.9 | 59.4 | 19.5 |
| MetaSSC (ours) | 80.3 | 69.7 | 59.5 | 20.7 |

Also, we have conducted more comparison experiments (Table 3) in another dataset, SemanticPOSS (Pan et al., 2020). SemanticPOSS is captured in dynamic and diverse campus environments. Compared to typical urban road scenarios, it offers more complex and variable scenes, including interactions between pedestrians, bicycles, and vehicles. The dataset features detailed semantic annotations, making it ideal for advancing 3D semantic understanding in real-world scenarios.

**Table 3.** Comparison with baseline models on the validation dataset of SemanticPOSS (Pan et al., 2020) benchmark

| Approach | Year | Input | SS | Semantic Scene Completion | | | | | | | | | | | |
|---|---|---|---|---|---|---|---|---|---|---|---|---|---|---|---|
| | | | IoU | mIoU | Person | Rider | Car | Trunk | Plants | Traffic-sign | Pole | Building | Fence | Bike | Ground |
| SSCNet | 2017 | $x^{pc}$ | 51.4 | 15.2. | 5.3 | 0.3 | 1.3 | 5.6 | 39.6 | 1.0 | 2.6 | 28.7 | 3.4 | 26.0 | 43.1 |
| LMSCNet | 2020 | $x^{pc}$ | 55.3 | 16.5 | 7.7 | 0.3 | 0.6 | 4.0 | 37.7 | 1.9 | 8.2 | 36.8 | 13.8 | 25.8 | 45.1 |
| MotionSC | 2020 | $x^{pc}$ | 52.7 | 17.6 | 7.8 | 0.5 | 0.5 | 3.9 | 39.8 | 2.2 | 8.5 | 39.2 | 13.1 | 30.8 | 47.0 |
| JS3CNet | 2021 | $x^{pc}$ | 58.1 | 22.7 | 18.9 | 0.2 | 7.1 | 3.6 | 47.8 | 2.2 | 0.0 | 46.3 | 26.6 | 43.4 | 53.2 |
| MetaSSC (ours) | - | $x^{pc}$ | 63.8 | 19.3 | 4.7 | 0.0 | 4.4 | 6.6 | 43.1 | 2.2 | 10 | 38.9 | 11.5 | 34.0 | 47.9 |

MetaSSC achieves the highest IoU of 63.8%, significantly outperforming baseline models, including JS3CNet at 58.1%. This result highlights the robustness of our method in understanding complex scenes, particularly in the challenging SemanticPOSS dataset. The high IoU demonstrates MetaSSC's effectiveness in capturing the overall structure of diverse environments.

Although the mIoU of MetaSSC is 19.4%, slightly lower than JS3CNet's 22.7%, the performance gap is largely attributable to the challenges in underrepresented or dynamic categories. Despite this, our method performs competitively across most individual categories. For instance, MetaSSC achieves strong results in static elements such as Traffic-sign (43.1%), Pole (22.2%), and Building (38.9%), which are crucial for accurate semantic understanding.

Dynamic elements like Bike (34.0%) and Car (4.4%) also show promising results, though some categories, such as Person (4.7%) and Fence (11.5%), present challenges due to their sparsity or occlusion. These observations underscore the potential for further optimization in handling rare or dynamic objects while maintaining strong overall scene comprehension.

The visual results in Fig. 5 showcase the performance of the models across various scenes in the SemanticKITTI validation dataset. These examples demonstrate how the proposed MetaSSC model can effectively handle complex urban scenes, with more accurate predictions for critical areas like roads and buildings.

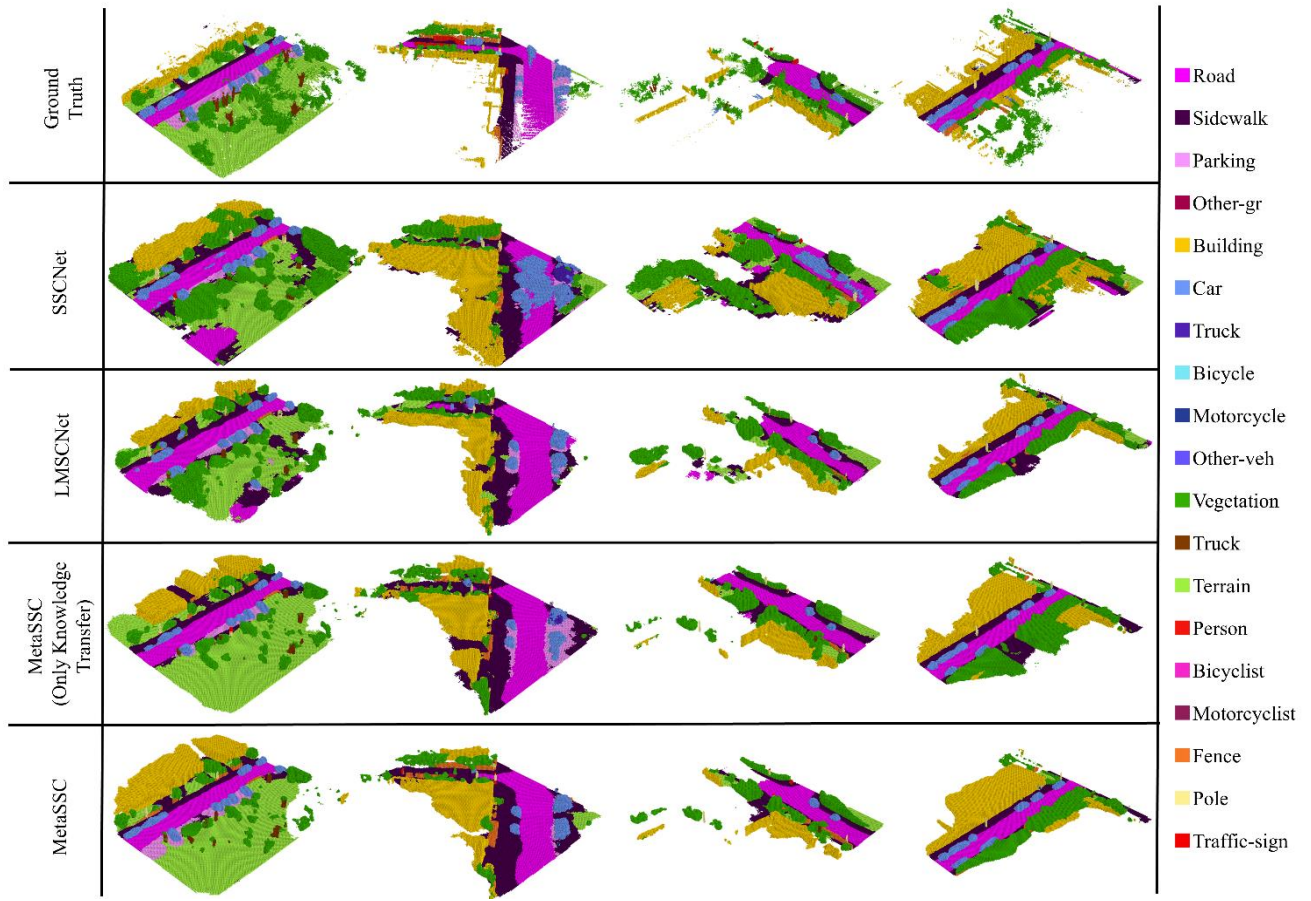

**Fig. 5.** Examples of SSC at full size on the SemanticKITTI validation dataset.

In summary, the combination of knowledge transfer and hierarchical multi-scale learning in MetaSSC allows the model to achieve a strong balance between accuracy and efficiency. While it leads in key metrics such as IoU and Precision, future work may focus on addressing class imbalance to further enhance recall and performance in underrepresented categories.

*4.2 Experiment 2: ablation analysis on critical techniques*

To further examine the contribution of critical techniques, an ablation analysis was conducted on the validation dataset of SemanticKITTI (Behley et al., 2019) benchmark. This analysis aims to isolate and assess the impact of key components of the proposed model by comparing different variant architectures. The four variant models, referred to as Multi-scaled, D-LKA, Transfer, and Mamba, are described as follows:

**(1) Multi-scaled**

LMSCNet serves as the base model for our analysis. It is a lightweight model that learns features at multiple resolutions, leveraging multi-scaled connections to capture both fine and broad contextual information. We incrementally improve our proposed method starting from this model to test how different components contribute to the final performance.

**(2) D-LKA**

In this variant, we replace the LMSCNet backbone with the deformable large-kernel attention network (Azad et al., 2024) to enhance feature extraction. This modification aims to improve the network's ability to predict complex 3D scenes more accurately.

**(3) Transfer**

This variant adopts the dual-phase training strategy discussed earlier to improve model performance and reduce training time. By pretraining on source datasets and fine-tuning on the target dataset, "Transfer" leverages knowledge from simulation domains to enhance real-world performance, ensuring faster convergence and improved generalization.

**(4) Mamba**

In this final variant, we integrate the Mamba block into the D-LKA network to handle the long-sequence modeling of 3D blocks. Mamba's strength lies in its ability to process sequential dependencies efficiently, which further enhances the model's understanding of 3D spatial structures for SSC.

The results of the ablation analysis are summarized in Table 2. As we progressed from "Multi-scaled" to "Mamba," the performance on all metrics either increased or remained consistent, with the exception of the Recall metric when D-LKA was introduced. The decrease in Recall during the D-LKA stage can be attributed to the trade-off between model complexity and generalization ability, as D-LKA focuses on learning richer features but may require more data for optimal recall. Overall, the results confirm that the techniques employed in our work are generally beneficial for SSC, showing consistent improvements across various performance metrics.

**Table 4.** Results of ablation experiments on the validation dataset of SemanticKITTI (Behley et al., 2019) benchmark.

| Approach | Precision | Recall | IoU | mIoU |
|---|---|---|---|---|
| Multi-scaled | 84.2 | 67.5 | 60.0 | 16.9 |
| D-LKA | 80.9 | 67.6 | 60.3 | 18.5 |
| Transfer | 83.5 | 74.0 | 64.7 | 20.9 |
| Mamba | 84.1 | 74.0 | 65.0 | 21.3 |

Additionally, we visualize the mIoU of the four models over training epochs on the validation dataset of SemanticKITTI in Fig. 6. "Multi-scaled" and "D-LKA" variants were trained directly on the target dataset, while "Transfer" and "Mamba" variants were pretrained on source datasets and fine-tuned on the target dataset. Notably, during the fine-tuning process, only the output layer was fine-tuned in the first epoch to stabilize early-stage training. The visualization clearly demonstrates that the dual-phase training strategy accelerates convergence and yields better performance across fewer epochs. This highlights the effectiveness of transferring pre-trained knowledge and fine-tuning on smaller target datasets to achieve desirable outcomes efficiently.

Considering the timeliness of calculations and computational capabilities, it is worth noting that our model has the longest training time, primarily due to the inclusion of the 3D DKLA module and Mamba. However, the proposed two-stage training method based on meta-learning is inherently model-agnostic. Typically, meta-pretraining is highly efficient, converging within just over an hour. In the adaptation phase, take Fig.6 as an example, a pretrained model can achieve excellent performance within just 10 epochs, whereas training a model from scratch cannot achieve similar performance even using 25 epochs. This demonstrates the efficiency and adaptability of our meta-learning framework.

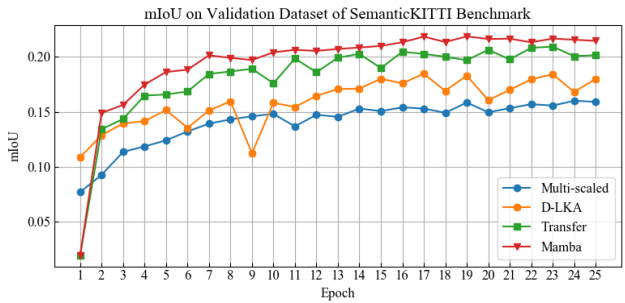

**Fig. 6.** mIoU on validation dataset of SemanticKITTI benchmark.

### *4.3 Experiment 3: few-shot capacity of meta-learning*

Since MetaSSC is a meta-learning-based method that facilitates knowledge transfer from source domains to the target domain, we investigated whether this kind of knowledge transfer could benefit the model in few-shot learning scenarios. Few-shot learning is critical in autonomous driving, where data collection is expensive and time-consuming, especially for less frequent events such as accidents or rare weather conditions. As part of this investigation, we also explored several techniques to assess their impact on the model's robustness under data scarcity conditions.

For these experiments, we used Scenario 00 and Scenario 01 of the training dataset from the SemanticKITTI (Behley et al., 2019) benchmark and evaluated the SSC metrics on the validation dataset. This setup allows us to simulate a data-scarce environment by limiting the training data to only two scenarios. Building on the D-LKA model introduced in the previous section, we extended the model with the following components: knowledge transfer, data augmentation on raw inputs, data augmentation on representations, and adversarial samples. Each of these components aims to either improve generalization under limited data or increase the robustness of the model to novel, unseen driving scenes.

The results of these experiments are summarized in Table 3, which explores the effectiveness of transfer learning, data augmentations, and adversarial samples on 3D SSC. All models were trained on the 0-th and 1-th scenarios of the train dataset and tested on the validation dataset of the SemanticKITTI (Behley et al., 2019) benchmark. The inclusion of transfer learning and data augmentation components provides notable improvements across several key classes, particularly those where the model traditionally struggled, such as "Sidewalk," "Car," and "Vegetation".

The results summarized in Table 3 show that knowledge transfer and data augmentations have a positive effect on model performance under data-scarcity scenarios. These techniques allow the model to generalize better by enriching the diversity of training data, either through augmented representations or knowledge transferred from other domains. While adding adversarial samples contributes only slightly to the overall performance, it improves results for specific classes, such as "Car" and "Vegetation." We hypothesize that adversarial samples help mitigate class imbalance to some extent by exposing the model to more challenging examples. However, this may come at the cost of disrupting the semantics and geometry of driving scenes, leading to reduced performance.

**Table 5.** Exploring the effectiveness of transfer learning, data augmentations and adversarial samples on 3D SSC.

| Approach | Road | Sidewalk | Parking | Other-gr | Building | Car | Truck | Bicycle | Motorcycle | Other-veh | Vegetation | Trunk | Terrain | Person | Bicyclist | Motorcyclist | Fence | Pole | Traffic-sign | mIoU |
|---|---|---|---|---|---|---|---|---|---|---|---|---|---|---|---|---|---|---|---|---|
| basic(30) | 59.9 | 32.1 | 8.8 | **0.0** | **35.8** | 39.2 | 0.2 | 0.1 | **0.7** | 5.0 | 38.9 | 0.2 | 44.3 | 0.0 | 0.0 | 0.0 | 8.7 | 20.6 | 6.2 | 17.1 |
| transfer(25) | 62.2 | 32.9 | 13.5 | 0.0 | 33.6 | 39.4 | 0.0 | 0.0 | 0.1 | 4.2 | 37.5 | 0.0 | 44.5 | 0.1 | **0.1** | 0.0 | 7.8 | 20.9 | 6.6 | 17.4 |
| transfer+outer(25) | 64.1 | 34.0 | 13.7 | 0.0 | 35.0 | 39.8 | 0.1 | 0.1 | 0.3 | 4.1 | 38.6 | 0.1 | **46.3** | 0.0 | 0.1 | 0.0 | **10.9** | **22.4** | 5.2 | **18.0** |
| transfer+inner(25) | 64.0 | 33.6 | **13.9** | 0.0 | 35.3 | 40.0 | 0.2 | 0.1 | 0.2 | 4.2 | 37.9 | 0.2 | 45.6 | 0.0 | 0.0 | 0.0 | 10.7 | 21.5 | 6.5 | 17.9 |
| transfer+adv(25) | **65.9** | **34.1** | 9.9 | 0.0 | 35.6 | **40.3** | **0.4** | **0.2** | 0.4 | 4.7 | 38.9 | **0.4** | 42.8 | **0.2** | 0.1 | 0.0 | 9.1 | 22.0 | **6.7** | 17.4 |

*Note: All models were trained on the 0-th and 1-th scenarios of train dataset, and tested on validation dataset of SemanticKITTI (Behley et al., 2019) benchmark.*

In conclusion, this ablation study demonstrates the effectiveness of incorporating knowledge transfer, data augmentation, and adversarial training into the MetaSSC for improving robustness in 3D SSC tasks, particularly under data-scarce conditions.

**Conclusions and discussions**

This work presents a meta-learning-based framework for tackling the SSC task in autonomous driving, focusing on efficiently transferring knowledge from simulations to real-world applications. By leveraging meta-knowledge acquired from simulated environments, the framework reduces the dependence on large-scale real-world data, which significantly lowers deployment costs and shortens development cycles. This approach not only enhances generalization to new environments but also ensures the model can adapt to diverse and dynamic driving scenarios.

A key innovation of this framework lies in its integration of LKA mechanisms and Mamba blocks within the backbone model. These components enable the model to effectively extract multi-scaled, long-sequence relationships from the sparse and irregular data provided by 3D voxel grids. LKA mechanisms allow the model to capture both local details and global context by expanding the receptive field without increasing computational complexity. In parallel, Mamba blocks improve the model's ability to process sequential dependencies across 3D blocks, enhancing the SSC task by capturing temporal and spatial relationships within driving scenes.

Our experimental results demonstrate that the proposed model has superior performance. The success of the dual-phase training strategy, which involves pretraining on simulated data followed by fine-tuning on real-world datasets, has been validated through extensive experiments. This strategy not only accelerates convergence but also improves the model's robustness and transferability to real-world conditions.

Additionally, we explored the model's capabilities under few-shot learning scenarios, which are critical for applications where data collection is limited or costly. Our findings indicate that knowledge transfer and data augmentations substantially improve performance in such settings. While the inclusion of adversarial samples offers only marginal improvements to the overall performance, it proves helpful in addressing class imbalance. This highlights the potential for further refinement in the use of adversarial techniques.

In summary, the combination of meta-learning, advanced attention mechanisms, and dual-phase training provides a scalable and robust solution for SSC in autonomous driving. The proposed framework not only improves the model's ability to handle complex and dynamic driving environments but also reduces deployment costs. These results pave the way for future advancements in SSC and contribute valuable insights for building safer and more reliable autonomous driving systems.

**Acknowledgement**

No acknowledgements.